\documentclass[11pt,a4paper]{article}
\usepackage[nohyperref]{naaclhlt2018}
\usepackage{times}
\usepackage{latexsym}
\usepackage{graphicx}
\usepackage{subcaption}
\usepackage{tikz}
\def\checkmark{\tikz\fill[scale=0.4](0,.35) -- (.25,0) -- (1,.7) -- (.25,.15) -- cycle;}

\newcommand{\dataset}{WinoBias}
\newcommand{\UW}{{\sf E2E}}
\newcommand{\Berkeley}{{\sf Feature}}
\newcommand{\Stanford}{{\sf Rule}}

\usepackage{array,multirow}
\graphicspath{figures}

\usepackage{url}

\aclfinalcopy % Uncomment this line for the final submission
\def\aclpaperid{1009} %  Enter the acl Paper ID here

\title{Gender Bias in Coreference Resolution:\\ Evaluation and Debiasing Methods}
\author{Jieyu Zhao$^\S$ \qquad
 Tianlu Wang$^\dagger$ \qquad 
  Mark Yatskar$^\ddag$   \\
{\bf Vicente Ordonez$^\dagger$ \qquad
 Kai-Wei Chang$^\S$}
\\
  $^\S$University of California, Los Angeles \qquad
  \{jyzhao, kwchang\}@cs.ucla.edu \\
  $^\dagger$ University of Virginia \qquad 
  \{tw8bc, vicente\}@virginia.edu 
  \\ $^\ddag$Allen Institute for Artificial Intelligence \qquad marky@allenai.org
}

\date{}

\begin{document}
\maketitle
\begin{abstract}
 %\jy{This paper mainly focus on detecting gender bias in multiple coreference systems: BerkeleyCoref, Stanford CoreNLP, UW end-to-end coreference system. We demonstarte that all of these coreference systems have gender bias issues through experiments. Furthermore, we propose a simple way, e.g., train a model on union data(original training data and the gender-reversed training data) to reduce gender bias.}

%Coreference resolution, a task consisting in clustering narrative phrases referring to the same entities, is considered fundamental in natural language processing. It requires reasoning that goes beyond grammar and syntax and has to take into account discourse, common-sense and semantic understanding. 

%Recent data-driven approaches have significantly advanced research in co-reference resolution but these systems carry the risk of inadvertently relying on societal stereotypes present in the data. 
We introduce a new benchmark, WinoBias, for coreference resolution focused on gender bias.
Our corpus contains Winograd-schema style sentences with entities corresponding to people referred by their occupation (e.g. the nurse, the doctor, the carpenter).
We demonstrate that a rule-based, a feature-rich, and a neural coreference system all link gendered pronouns to pro-stereotypical entities with higher accuracy than anti-stereotypical entities, by an average difference of 21.1 in F1 score.
%and is perfectly balanced in terms of gendered references to each entity (e.g. he, she, her, him).
%We show that current co-reference resolution systems perform significantly worse in the presence of gender atypical roles.
%(as defined using labor demographic statistics). % KW: can move this detail to intro or method section.
Finally, we demonstrate a data-augmentation approach that, in combination with existing word-embedding debiasing techniques, removes the bias demonstrated by these systems in WinoBias without significantly affecting their performance on existing coreference benchmark datasets.
Our dataset and code are avialable at \url{http://winobias.org}.

%Furthermore, we show that simple baselines that apply debiasing to the word embeddings used in some of these systems, along with gender anonymization, and data augmentation techniques, achieve promising results in alleviating this problem. %We systematically observe that given the same context, coreference systems make significantly different predictions for \texttt{female} and \texttt{male} entities. 
%For example, they cannot identify that a pronoun ``she'' refers to ``the leader'' as such a co-referent pair seldom appears on the training data.
%In this paper, we analyze the impact of gender bias in a coreference resolution corpus and three coreference resolution systems. We propose an approach to reduce gender stereotypes inadvertently encoded in these coreference systems. Our proposed method is easy to implement and does not need any modification to the training procedure. Experiments show that the proposed approach effectively reduce gender bias without affecting the accuracy of the original system. 
\end{abstract}
\section{Introduction}
\label{sec:introduction}

\begin{figure}[t]
    \centering
        \includegraphics[width=\linewidth]{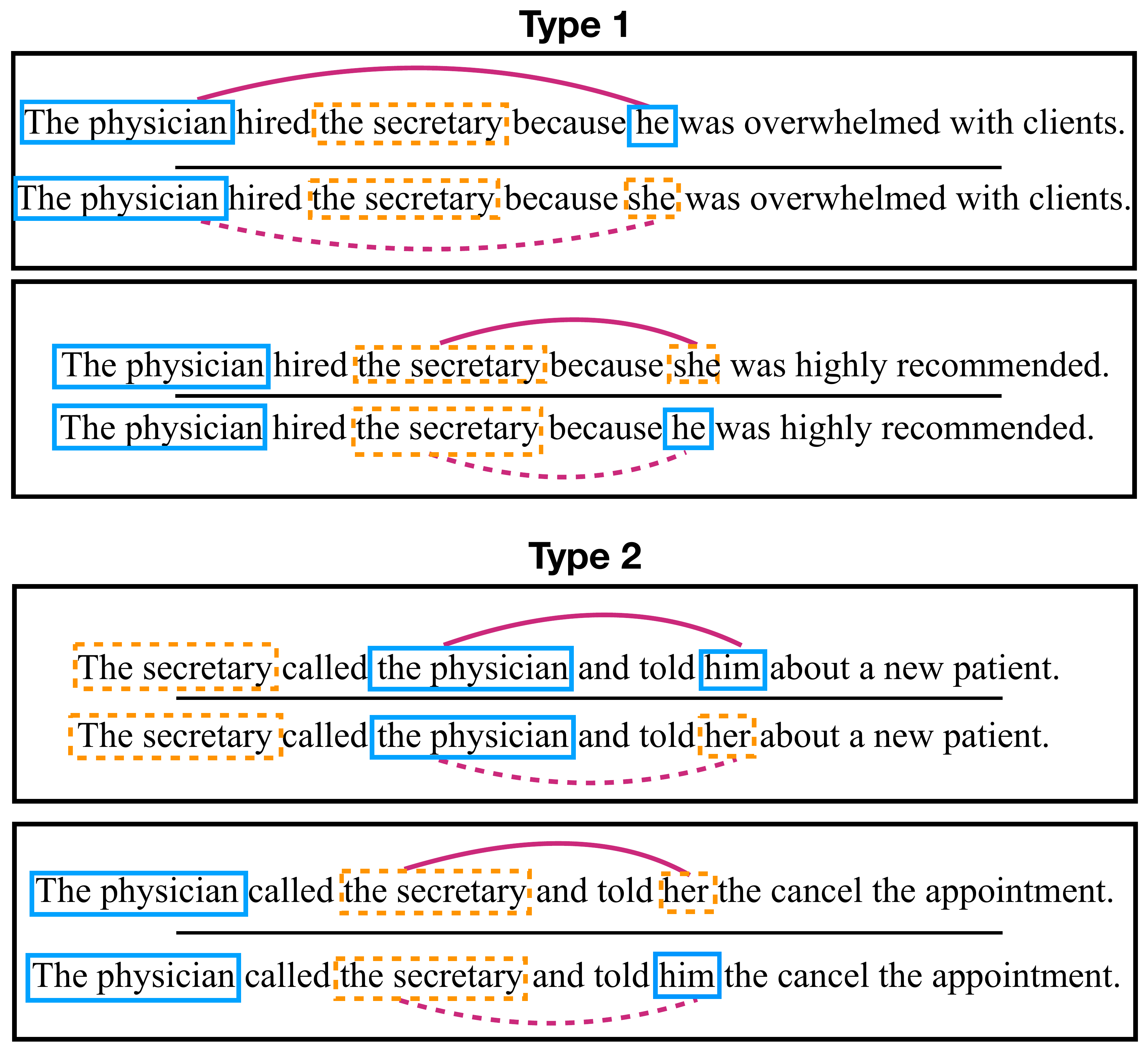}
        \vspace{-20pt}
        \label{fig:president_he}
    \caption{\small Pairs of gender balanced co-reference tests in the~\dataset~dataset. Male and female entities are marked in solid blue and dashed orange, respectively. For each example, the gender of the pronominal reference is irrelevant for the co-reference decision. Systems must be able to make correct linking predictions in pro-stereotypical scenarios (solid purple lines) and anti-stereotypical scenarios (dashed purple lines) equally well to pass the test. Importantly, stereotypical occupations are considered based on US Department of Labor statistics.}
    \vspace{-10pt}
\label{fig:teaser}
\end{figure}

Coreference resolution is a task aimed at identifying phrases (mentions) referring to the same entity. %, has been a fundamental task in NLP. % and has been adopted in a wide range of applications, including  document summarization, question answering and information extraction~\cite{bengtson2008understanding,lee2011stanford}.
Various approaches, including rule-based~\cite{raghunathan2010multi}, feature-based~\cite{DurrettKlein2013,PengChRo15}, and neural-network based ~\cite{clark2016deep,lee2017end} have been proposed. 
While significant advances have been made, systems carry the risk of relying on societal stereotypes present in training data that could significantly impact their performance for some demographic groups.

In this work, we test the hypothesis that co-reference systems exhibit gender bias by creating a new challenge corpus, \dataset.%\footnote{{https://uclanlp.github.io/corefBias/}}
This dataset follows the winograd format~\cite{hirst,RahmanNg12c,peng2015solving}, and contains references to people using a vocabulary of 40 occupations.
It contains two types of challenge sentences that require linking gendered pronouns to either male or female stereotypical occupations (see the illustrative examples in Figure~\ref{fig:teaser}).
None of the examples can be disambiguated by the gender of the pronoun but this cue can potentially distract the model. 
%and each type contains both stereotype and non\_stereotype subset.
We consider a system to be gender biased if it links pronouns to occupations dominated by the gender of the pronoun (pro-stereotyped condition) more accurately than occupations not dominated by the gender of the pronoun (anti-stereotyped condition).
%\my{I want to add a sentence here about that we are providing a weak certification, but it sounds very negative here Maybe : Our corpus can be used to certify a system has gender bias but we are unable to certify systems are gender bias free. (my : I like that lot) How about: }
The corpus can be used to certify a system has gender bias.\footnote{Note that the counter argument (i.e., systems are gender bias free) may not hold.}
%Such a test provides a weak guarantee 
%The dataset is large, containing over 3000 sentences.

We use three different systems as prototypical examples: the Stanford Deterministic Coreference System~\cite{raghunathan2010multi}, the Berkeley Coreference Resolution System~\cite{DurrettKlein2013} and the current best published system: the UW End-to-end Neural Coreference Resolution System~\cite{lee2017end}.
Despite qualitatively different approaches, all systems exhibit gender bias, showing an average difference in performance between pro-stereotypical and anti-stereotyped conditions of $21.1$ in F1 score.
Finally we show that given sufficiently strong alternative cues, systems can ignore their bias. 
%\jy{Concurrent work~\cite{Rachel18} also shows bias in coreference resolution systems.}

%it should provide similar results on these two subsets. However, the aforementioned models all show great difference on the stereotype and non\_stereotype datasets which also prove the potential gender bias issue.
In order to study the source of this bias, we analyze the training corpus used by these systems, Ontonotes 5.0~\cite{weischedel2012ontonotes}.\footnote{The corpus is used in CoNLL-2011 and CoNLL-2012 shared tasks, http://www.conll.org/previous-tasks} Our analysis shows that female entities are significantly underrepresented in this corpus. % and it exhibits strong gender biases.
To reduce the impact of such dataset bias, we propose to generate an auxiliary dataset where all male entities are replaced by female entities, and vice versa, using a rule-based approach.
Methods can then be trained on the union of the original and auxiliary dataset.
In combination with methods that remove bias from fixed resources such as word embeddings~\cite{BCWS16}, our data augmentation approach completely eliminates bias when evaluating on \dataset~, without significantly affecting overall coreference accuracy.

\section{\dataset}
\label{sec:calibrating_bias}

%\subsection{Winograd Data Creation}
To better identify gender bias in coreference resolution systems, we build a new dataset centered on people entities referred by their occupations from a vocabulary of 40 occupations gathered from the US Department of Labor, shown in Table~\ref{tab:gender_table}.\footnote{Labor Force Statistics from the Current Population Survey, 2017. https://www.bls.gov/cps/cpsaat11.htm} 
We use the associated occupation statistics to determine what constitutes gender stereotypical roles (e.g. 90\% of nurses are women in this survey). 
Entities referred by different occupations are paired and used to construct test case scenarios.
%All co-reference decisions should be resolved without gender.
Sentences are duplicated using male and female pronouns, and contain equal numbers of correct co-reference decisions for all occupations.
In total, the dataset contains 3,160 sentences, split equally for development and test, created by researchers  familiar with the project.
%Each sentence in the dataset contains a pair of occupations (X and Y), one is male biased, the other is female biased. It  also contains one gender pronoun. 
%The dataset contains two types of sentences containing gendered pronoun references to one of the male or female stereotyped professions. 
Sentences were created to follow two prototypical templates but annotators were encouraged to come up with scenarios where entities could be interacting in plausible ways. %\footnote{For example, nurses treating construction workers for injuries.} 
Templates were selected to be challenging and designed to cover cases requiring semantics and syntax separately.\footnote{We do not claim this set of templates is complete, but that they provide representative examples that, pratically, show bias in existing systems.} 

%We considered two types of prototypical test scenarios based on rough templates:

%\vspace{-5pt}
\paragraph{Type 1: [\texttt{entity1}] [interacts with]   [\texttt{entity2}] [conjunction] [pronoun] [circumstances].} Prototypical WinoCoRef style sentences, where co-reference decisions must be made using world knowledge about given circumstances (Figure~\ref{fig:teaser}; Type 1).
Such examples are challenging because they contain no syntactic cues.

%\vspace{-5pt}
\paragraph{Type 2: [\texttt{entity1}] [interacts with]  [\texttt{entity2}] and then [interacts with] [pronoun] for [circumstances].} These tests can be resolved using syntactic information and understanding of the pronoun (Figure~\ref{fig:teaser}; Type 2). 
We expect systems to do well on such cases because both semantic and syntactic cues help disambiguation.

\begin{table}[t]
    \small
    \centering
    \begin{tabular}{|l|r|l|r|}
    \hline
    Occupation & \% & Occupation & \% \\
    \hline

carpenter &2 &  editor  &52  \\ 
mechanician &  4 & designers &54  \\
construction worker & 4 &  accountant  &61  \\ 
laborer & 4  & auditor &   61  \\
driver  & 6 & writer  &63  \\  
sheriff &14 & baker &65  \\ 
mover & 18   & clerk &72  \\ 
developer &20 &   cashier & 73  \\  
farmer  &22 &   counselors  &73  \\  
guard &22 &    attendant &  76  \\
chief & 27   & teacher & 78  \\ 
janitor & 34    &  sewer &80  \\
lawyer  &35 &     librarian &84  \\
cook &  38   & assistant & 85  \\
physician&  38 &   cleaner & 89  \\ 
ceo &39 &    housekeeper &89  \\
analyst &41 &    nurse & 90  \\
manager &43 &    receptionist  &90  \\
supervisor & 44   &  hairdressers  &92  \\
salesperson&  48 &   secretary & 95  \\
\hline
    \end{tabular}
    \caption{\small Occupations statistics used in WinoBias dataset, organized by the percent of people in the occupation who are reported as female. When woman dominate profession, we call linking the noun phrase referring to the job with female and male pronoun as `pro-stereotypical`, and `anti-stereotypical`, respectively. Similarly, if the occupation is male dominated, linking the noun phrase with the male and female pronoun is called, `pro-stereotypical` and `anti-steretypical`, respectively.}
    \label{tab:gender_table}
\end{table}
%one follows the routine ``X verb Y pronoun disambiguation''.  The other one is of the form: ``X verb1 Y and verb2 pronoun''. The second type of sentence is syntactically unambiguous. Considering the following examples:\\
%Type1: \textit{[The developer]$_{e_1}$ was unable to communicate with [the writer]$_{e_2}$ because [he]$_{pro\_1}$ only understands code.} \\
%Type2: \textit{[The writer]$_{e_1}$ handed [the developer]$_{e_2}$ a coffee and then smiled at [him]$_{pro\_2}$.}

%All the sentences cannot be resolved by just looking at the gender or the pronoun.

%Besides the two types, we also split the dataset to stereotype and non\_stereotype parts. The first one contains the sentences that following the statistical stereotype in professions, for instance, it will use ``she'' to refer to  ``the writer'' and ``he'' to refer to ``the developer''. The non\_stereotype dataset instead will use ``he'' to stand for ``the writer'' and ``she'' to mention ``the developer''.

%We evaluate the three systems on this new dataset.
%\vspace{-5pt}
\paragraph{Evaluation}
To evaluate models, we split the data in two sections: one where correct co-reference decisions require linking a gendered pronoun to an occupation stereotypically associated with the gender of the pronoun and one that requires linking to the anti-stereotypical occupation.
%If a model is gender biased, it can .
We say that a model passes the WinoBias test if for both Type 1 and Type 2 examples, pro-stereotyped and anti-stereotyped co-reference decisions are made with the same accuracy.
%A limitation of this test is that it can be used to prove a system is biased but not that it is unbiased.

%The results  in Table~\ref{tab:cluster_gender_ratio} shows that all the three systems perform better on the stereotyped dataset than on the non\_stereotyped dataset for both two types which means all of them have the  gender bias issue. The rule-based Stanford system is the most gender biased among the three systems. And UW system is more gender biased than the Berkeley system. The Stanford and Berkeley systems both rely on the gender file collected by~\cite{Bergsma:06}, but the Berkeley system will also rely on other features which makes it less biased than the Stanford System. However the UW system is the end-to-end system. All the information is collected from the dataset itself, which makes it more sensitive to the dataset.

\section{Gender Bias in Co-reference}
\label{sec:visualizing_bias}

\begin{table*}[t]
\small
\centering
    \begin{tabular}{|c|c|c|c||c||c|c|c|c||c|c|c|c|}
 
        \hline
          Method & Anon. & Resour. &  Aug. & OntoNotes & T1-p &  T1-a & Avg & $\mid$ Diff $\mid$ & T2-p & T2-a & Avg & $\mid$ Diff $\mid$  \\
          \hline
           \UW &&&&{\bf 67.7}&{\bf 76.0}&49.4& 62.7&26.6* 
           &{\bf 88.7}&75.2 & 82.0 &13.5*\\
           \UW & \checkmark &&&66.4&73.5 & 51.2 &62.6&21.3*
           & 86.3 & 70.3&78.3&16.1* \\
           \UW & \checkmark & \checkmark &&66.5 &67.2&59.3&{63.2}&7.9*
           &81.4&82.3&81.9&0.9 \\
           \UW &\checkmark  & & \checkmark &66.2&65.1&59.2&62.2&5.9*
           &86.5&{\bf 83.7}&{\bf 85.1}& 2.8* \\
           \UW & \checkmark & \checkmark & \checkmark &66.3 &63.9&{\bf 62.8}&{\bf 63.4}&{\bf 1.1}
           &81.3&{ 83.4}&82.4&{\bf 2.1} \\
          \hline
          \hline
           \Berkeley &&&&{\bf 61.7}& {\bf 66.7} &56.0&{61.4}&10.6*&{\bf 73.0}&57.4&65.2&15.7* \\
           \Berkeley & \checkmark &&&61.3 &65.9&56.8&61.3&9.1*&72.0&58.5&65.3&13.5* \\
           \Berkeley & \checkmark & \checkmark &&61.2 &61.8&{\bf 62.0}&{\bf 61.9}&{\bf 0.2}&67.1&63.5&65.3&3.6 \\
           \Berkeley & \checkmark &  &\checkmark& 61.0&65.0&{57.3}&61.2&{7.7*}&{ 72.8}&{ 63.2}&{68.0}&{9.6*} \\
              \Berkeley & \checkmark & \checkmark &\checkmark& 61.0 &62.3&{60.4}&61.4&{1.9*}&{71.1}&{\bf 68.6}&{\bf 69.9}&{\bf 2.5} \\
          \hline
          \hline
           \Stanford &&&&57.0&76.7&37.5&57.1 & 39.2* &50.5 & 29.2& 39.9 & 21.3*\\
           
         % 10 & Stanford & \checkmark &&&&& \\
         
        \hline
    \end{tabular}
    \vspace{-10pt}

    \caption{\small F1 on OntoNotes and WinoBias development set. WinoBias results are split between Type-1 and Type-2 and in pro/anti-stereotypical conditions. * indicates the difference between pro/anti stereotypical conditions is significant ($p < .05$) under an approximate randomized test~\cite{graham2014randomized}. Our methods eliminate the difference between pro-stereotypical and anti-stereotypical conditions (Diff), with little loss in performance (OntoNotes and Avg). }
    \label{tab:results}
   % \vspace{-10pt}

\end{table*}

\begin{table*}[t]
\small
\centering
    \begin{tabular}{|c|c|c|c||c||c|c|c|c||c|c|c|c|}
 
        \hline
          Method & Anon. & Resour. &  Aug. & OntoNotes & T1-p &  T1-a & Avg & $\mid$ Diff $\mid$ & T2-p & T2-a & Avg & $\mid$ Diff $\mid$  \\
          \hline
           \UW &&&&{\bf 67.2}&{\bf 74.9}&47.7&61.3&27.2* 
           &{\bf 88.6}&77.3 & {\bf 82.9} &11.3*\\
           \UW & \checkmark & \checkmark & \checkmark &66.5 &62.4&{\bf 60.3}&{\bf 61.3}&{\bf 2.1}
           &78.4&{\bf 78.0}& 78.2 &{\bf 0.4} \\
          \hline
          \hline
           \Berkeley &&&&{\bf 64.0}& {\bf 62.9} &58.3&{60.6}&4.6*&{\bf 68.5}&57.8& 63.1 & 10.7* \\
           \Berkeley & \checkmark & \checkmark &\checkmark& 63.6 &62.2&{\bf 60.6}& {\bf 61.4}& {\bf 1.7} &{\bf 70.0}&{\bf 69.5}&{\bf 69.7}&{\bf 0.6} \\
          \hline
          \hline
           \Stanford &&&&58.7&72.0&37.5& 54.8 & 34.5* & 47.8 & 26.6 & 37.2 & 21.2*\\
           
         % 10 & Stanford & \checkmark &&&&& \\
         
        \hline
    \end{tabular}
    \vspace{-10pt}

    \caption{\small F1 on OntoNotes and Winobias test sets. Methods were run once, supporting development set conclusions.}
    \label{tab:test_results}
\vspace{-10pt}
\end{table*}

%%KW: Talk about the new dataset we collect and provide data statistics

In this section, we highlight two sources of gender bias in co-reference systems that can cause them to fail \dataset: training data and auxiliary resources and propose strategies to mitigate them.
%We first, look at bias coming directly from training data.
%Then we consider bias from auxiliary resources.

%analyze the coreference resolution corpus and existing systems and show that both the corpus and the models are biased. 
%To simplify the analysis, we consider only the binary gender in this work.
%However, the proposed approach can be applied to other types of biases. 
%Besides this, we propose a new winograd dataset which can be used as a benchmark to analyze the gender bias issue in the coreference systems.

\subsection{Training Data Bias}
\label{sub:ccb}

\paragraph{Bias in OntoNotes 5.0}
Resources supporting the training of co-reference systems have severe gender imbalance. 
In general, entities that have a mention headed by gendered pronouns (e.g.``he'', ``she'') are over 80\% male.\footnote{To exclude mentions such as ``his mother'', we use Collins head finder~\cite{collins2003head} to identify the head word of each mention, and only consider the mentions whose head word is gender pronoun.}
Furthermore, the way in which such entities are referred to, varies significantly. 
Male gendered mentions are more than twice as likely to contain a job title as female mentions.\footnote{We pick more than  900 job titles from a gazetteer. %\url{https://sourceforge.net/p/gate/code/HEAD/tree/gate/trunk/plugins/ANNIE/resources/gazetteer/jobtitles.lst}
}
Moreover, these trends hold across genres.

\vspace{-5pt}
\paragraph{Gender Swapping}
To remove such bias, we construct an additional training corpus where all male entities are swapped for female entities and vice-versa.
Methods can then be trained on both original and swapped corpora.
%in the OntoNotes corpus.
This approach maintains non-gender-revealing correlations while eliminating correlations between gender and co-reference cues.

We adopt a simple rule based approach for gender swapping. 
First, we anonymize named entities using an automatic named entity finder~\cite{lample2016neural}. Named entities are replaced consistently within document (i.e. ``Barak Obama ... Obama was re-elected.'' would be annoymized to ``E1 E2 ... E2 was re-elected.'' ).
Then we build a dictionary of gendered terms and their realization as the opposite gender by asking workers on Amazon Mechnical Turk to annotate all unique spans in the OntoNotes development set.\footnote{Five turkers were presented with anonymized spans and asked to mark if it indicated male, female, or neither, and if male or female, rewrite it so it refers to the other gender.}
Rules were then mined by computing the word difference between initial and edited spans.
Common rules included ``she $\rightarrow$ he'', ``Mr.'' $\rightarrow$ ``Mrs.'', ``mother'' $\rightarrow$ ``father.''
Sometimes the same initial word was edited to multiple different phrases: these were resolved by taking the most frequent phrase, with the exception of ``her $\rightarrow$ him'' and ``her $\rightarrow$ his'' which were resolved using part-of-speech.
Rules were applied to all matching tokens in the OntoNotes. We maintain anonymization so that cases like ``John went to his house'' can be accurately swapped to ``E1 went to her house.''
\subsection{Resource Bias}
%\vspace{-5pt}
\paragraph{Word Embeddings}
Word embeddings are widely used in NLP applications however recent work has shown that they are severely biased: ``man'' tends to be closer to ``programmer'' than ``woman''~\cite{BCWS16,CBN17}.
Current state-of-art co-reference systems build on word embeddings and risk inheriting their bias. %some co-ref systems that also use word embeddings.
To reduce bias from this resource, we replace GloVe embeddings with debiased vectors~\cite{BCWS16}.
%When the coreference resolution system is built on such word embeddings, it will  also be influenced by such stereotype.
\vspace{-5pt}
\paragraph{Gender Lists}
While current neural approaches rely heavily on pre-trained word embeddings, previous feature rich and rule-based approaches rely on corpus based gender statistics mined from external resources~\cite{Bergsma:06}.
Such lists were generated from large unlabeled corpora using heuristic data mining methods. %. and have been provided with co-reference training data since \my{XXX}. 
 These resources provide counts for how often a noun phrase is observed in a male, female, neutral, and plural context.
To reduce this bias, we balance male and female counts for all noun phrases.

\section{Results}
\label{sec:results}

 \begin{table}[t]
 \small
 \centering
 \begin{tabular}{|l|c|c|}
         \hline 
          Model & Original & Gender-reversed \\
         \hline
         E2E & 66.4 &65.9 \\
         \hline
         Feature & 61.3 & 60.3\\
         \hline
         %Rule & 56.3  & 55.8 \\
         %\hline
     \end{tabular}
     \caption{ \small Performance on the original and the gender-reversed developments dataset (anonymized).}
     \label{tab:ontonotes_swapped}
 \end{table}

%In this section we present evaluation of three representative co-reference systems on both  OntoNotes 5.0 and ~\dataset~.

%We present evaluations on both OntoNotes 5.0 and on ~\dataset~.   
 %We have shown the gender bias issue in the coreference systems. 
 %In this section, we are going to propose methods to reduce the gender bias.
% What we do here is that we have the access to the training dataset and we will try to fix the problem by just adding some redundant information based on the original training dataset. 
%The first method is to add some redundant information based on the original training dataset. Without changing any training procedure or losing any useful information, we can show that this algorithm  can help to reduce the gender bias in the coreference systems.
%As the word embeddings also shows implicit gender bias issue~\cite{BCWS16}, the coreference systems using the word embeddings will also inherit such problem.  Besides the data augmentation method, we also  incorporate the debiased word embeddings for the calibration.

%\subsection{Results}
In this section we evaluate of three representative systems: 
rule based, \Stanford ,~\cite{raghunathan2010multi}, feature-rich, \Berkeley, ~\cite{DurrettKlein2013}, and end-to-end neural (the current state-of-the-art), \UW, ~\cite{lee2017end}.
The following sections show that performance on \dataset~reveals gender bias in all systems, that our methods remove such bias, and that systems are less biased on OntoNotes data. 

\paragraph{WinoBias Reveals Gender Bias}
Table~\ref{tab:results} summarizes development set evaluations using all three systems.
Systems were evaluated on both types of sentences in~\dataset~(T1 and T2), separately in pro-stereotyped and anti-stereotyped conditions ( T1-p vs. T1-a, T2-p vs T2-a).
We evaluate the effect of named-entity anonymization (Anon.), debiasing supporting resources\footnote{Word embeddings for \UW~and gender lists for \Berkeley} (Resour.) and using data-augmentation through gender swapping (Aug.).
\UW~and \Berkeley~were retrained in each condition using default hyper-parameters while \Stanford~was not debiased because it is untrainable. 
%Debiasing techniques were not applied to the rule based system because it is not trainable.
We evaluate using the coreference scorer v8.01~\cite{pradhan2014scoring} and compute the average (Avg) and absolute difference (Diff) between pro-stereotyped and anti-stereotyped conditions in \dataset.

All initial systems demonstrate severe disparity between pro-stereotyped and anti-stereotyped conditions. 
Overall, the rule based system is most biased, followed by the neural approach and feature rich approach. 
Across all conditions, anonymization impacts \UW~ the most, while all other debiasing methods result in insignificant loss in performance on the OntoNotes dataset.
Removing biased resources and data-augmentation reduce bias independently and more so in combination, allowing both \UW~and \Berkeley~to pass \dataset~ without significantly impacting performance on either OntoNotes or \dataset~. 
Qualitatively, the neural system is easiest to de-bias and %, requiring little knowledge of it's inner workings.
our approaches could be applied to future end-to-end systems.
Systems were evaluated once on test sets, Table~\ref{tab:test_results}, supporting our conclusions.

\paragraph{Systems Demonstrate Less Bias on OntoNotes} 
While we have demonstrated co-reference systems have severe bias as measured in \dataset~, this is an out-of-domain test for systems trained on OntoNotes.
Evaluating directly within OntoNotes is challenging because sub-sampling documents with more female entities would leave very few evaluation data points.
Instead, we apply our gender swapping system (Section \ref{sec:visualizing_bias}), to the OntoNotes development set and compare system performance between swapped and unswapped data.\footnote{This test provides a lower bound on OntoNotes bias because some mistakes can result from errors introduce by the gender swapping system.}
If a system shows significant difference between original and gender-reversed conditions, then we would consider it gender biased on OntoNotes data.

Table~\ref{tab:ontonotes_swapped} summarizes our results.
The \UW~system does not demonstrate significant degradation in performance, while \Berkeley~loses roughly 1.0-F1.\footnote{We do not evaluate the \Stanford~ system as it cannot be train for anonymized input.}
This demonstrates that given sufficient alternative signal, systems often do ignore gender biased cues.
%For example, in , the models do not show bias on in-domain OntoNotes data which is indicated by the similar performance on the gender reversed dataset as  sufficient cues are covered in the dataset.
On the other hand, \dataset~provides an analysis of system bias in an adversarial setup, showing, when examples are challenging, systems are likely to make gender biased predictions.

%While systems are less accurate in the gender-reversed setting, the effect is small, likely because either ambiguous situations are infrequent or there is sufficient alternative signal.

%word vectors and gender lists 

%We show the performance of the three coreference resolution systems before and after retraining using our calibration in Table~\ref{tab:cluster_gender_ratio}.
%Before adopting our calibration method, the UW system shows $21\%$ difference on the type1 wino dataset.
%After the ART method, the difference shrinks to $5.95$. Also, with the DRT method, we can  reduce the gap to $7.94\%$ which means both method can help to mitigate the gender bias issue in the system. However the DART method, which makes the difference to $1.1\%$, shows the best performance in reducing the gender bias.

%To make the results more straightforward, we only evaluate the models on the subset that we have changed the gender. The results are shown in Table~\ref{tab:changed_only}.
%All the systems achieve sub-optimal performance on the gender-reversed corpus.
%The performance of Stanford CoreNLP system drops $1.6\%$. The performance of Berkeley and UW systems also drops $1.46\%$, $1.23\%$, respectively. 
%It indicates that the model is sensitive to the gender of entities and treats male and female entities differently which is mainly due to the model fails to identify entities that are referred by female pronouns in the gender-reversed corpus.

\section{Related Work}
%\paragraph{Fairness in Machine Learning}
Machine learning methods are designed to generalize from observation but if algorithms inadvertently learn to make predictions based on stereotyped associations they risk amplifying existing social problems.
%
%Machine learning models are often built so that they generalize observations from the data they are trained on. 
%For example, a model learns to identify a cat in an image by looking at the common characteristics of the subject. 
%Generalization allows models to make predictions on unseen instances at test time. 
%However, when such generalization ability is applied to variables such as gender, age, or race, there is a risk of making biased decisions based on general impressions of a demographic group.
Several problematic instances have been demonstrated, for example, word embeddings can encode sexist stereotypes~\cite{BCWS16,CBN17}.
%For instance, when an embedding model is used to solve an analogy puzzle such as ``man is to woman as surgeon to who'', the model responds with ``nurse'', as it learns that male is more associated with surgeon and female is more closely associated with nurse.
Similar observations have been made in vision and language models~\cite{ZWYOC17}, online news~\cite{ross2011women}, web search~\cite{kay2015unequal} and advertisements~\cite{sweeney2013discrimination}. %also show issues with gender bias.
In our work, we add a unique focus on co-reference, and propose simple general purpose methods for reducing bias.

Implicit human bias can come from imbalanced datasets.
When making decisions on such datasets, it is usual that under-represented samples in the data are neglected since they do not influence the overall accuracy as much.
For binary classification~\citet{kamishima2012fairness,kamishima2011fairness} add a regularization term to their objective that penalizes biased predictions. 
Various other approaches have been proposed to produce ``fair'' classifiers~\cite{calders2009building,feldman2015certifying,misra2016seeing}. 
For structured prediction, the work of~\citet{ZWYOC17} reduces bias by using corpus level constraints, but is only practical for models with specialized structure.
\citet{kusner2017counterfactual} propose the method based on  causal inference to achieve the model fairness where they do the data augmentation under specific cases, however,  to the best of our knowledge, we are the first to propose data augmentation based on gender swapping in order to reduce gender bias. 

Concurrent work~\cite{Rachel18} also studied gender bias in coreference resolution systems, and created a similar job title based, winograd-style, co-reference dataset to demonstrate bias~\footnote{Their dataset also includes gender neutral pronouns and examples containing one job title instead of two.}. 
Their work corroborates our findings of bias and expands the set of systems shown to be biased while we add a focus on debiasing methods.
Future work can evaluate on both datasets.

\section{Conclusion}
\label{sec:conclusion}

Bias in NLP systems has the potential to not only mimic but also amplify stereotypes in society. 
For a prototypical problem, coreference, we provide a method for detecting such bias and show that three systems are significantly gender biased.
We also provide evidence that systems, given sufficient cues, can ignore their bias.
Finally, we present general purpose methods for making co-reference models more robust to spurious, gender-biased cues while not incurring significant penalties on their performance on benchmark datasets. 

%In this work, the corpus we use and the dataset we build is in English. Also here we consider the binary gender case. In the future it will be of much interest to work on some other languages such as Turkish where we cannot tell the gender from the singular third person pronouns. 

%We show that in three prototypical co-reference systems, gender bias from both supporting resources and training data significantly impact system performance, and 

%\my{impact statement here, because well, we have no space anywhere else...}
%To the best of our knowledge, our work is the first to demonstrate gender bias issues in coreference resolution and  propose a new dataset to verify this bias.
%The proposed algorithms are simple but effective.
%We are interested in extending this work by applying our methods to other tasks.%artificial intelligence tasks and designing new algorithms to mitigate the impact of bias in data by adding regularization in the training objective function.
% propose an approach to reduce this bias by generating a gender-reversed corpus. The proposed ART algorithm is simple but effective and does not require additional data annotations. We are interested in extending this work by applying ART to other artificial intelligence tasks and designing new algorithms to mitigate the impact of bias in data by adding regularization in the training objective function. 

\section*{Acknowledgement}
This work was supported in part by National Science Foundation Grant IIS-1760523, two NVIDIA GPU Grants, and a Google Faculty Research Award. We would like to thank Luke Zettlemoyer, Eunsol Choi, and Mohit Iyyer for helpful discussion and feedback.

%\bibliography{naaclhlt2018}
\bibliography{naaclhlt2018}
\bibliographystyle{acl_natbib}

% \appendix
% \section{Appendix}
% \input{sections/appendix.tex}
% \label{sec:supplemental}

% \section{Multiple Appendices}
% \dots can be gotten by using more than one section. We hope you won't
% need that.

\end{document}